%% file: main.tex
\documentclass{article}
\ifdefined\pdfsuppressptexinfo\pdfsuppressptexinfo=-1\fi
\ifdefined\pdftrailerid\pdftrailerid{}\fi

\usepackage[T1]{fontenc}
\usepackage{microtype}
\usepackage{graphicx}
\usepackage{booktabs}
\usepackage{hyperref}
\usepackage{xurl}
\usepackage[preprint]{icml2027}
\makeatletter
\renewcommand{\ICML@preprint}{\textit{Preprint. November 2026.}}
\makeatother

\usepackage{amsmath}
\usepackage{amssymb}
\usepackage{xcolor}
\usepackage[capitalize,noabbrev]{cleveref}

\input{numbers.tex}

\newcommand{\pp}{\%}

\icmltitlerunning{Exact on Paper: Serving Shortcuts and the Graded Answer}

\hypersetup{pdftitle={Exact on Paper: Which LLM Serving Shortcuts Flip Graded Answers, Measured Against the Noise of Reordering the Same Requests}, pdfauthor={Sharven Rane}, pdfsubject={Preprint}, pdfkeywords={LLM serving, vLLM, speculative decoding, prefix caching, chunked prefill, KV cache quantization, batch invariance, nondeterminism, release gating}, pdfcreator={LaTeX}}
\begin{document}

\twocolumn[
  \icmltitle{Exact on Paper: Which LLM Serving Shortcuts Flip Graded Answers,\\
    Measured Against the Noise of Reordering the Same Requests}
  \begin{icmlauthorlist}
    \icmlauthor{Sharven Rane}{ind}
  \end{icmlauthorlist}
  \icmlaffiliation{ind}{Independent Researcher}
  \icmlcorrespondingauthor{Sharven Rane}{sharvenrane7@gmail.com}
  \icmlkeywords{LLM serving, vLLM, speculative decoding, prefix caching, chunked prefill, KV cache quantization, batch invariance, nondeterminism, release gating}
  \vskip 0.3in
]

\printAffiliationsAndNotice{}

\begin{abstract}
Teams enable serving features such as prefix caching, chunked prefill and speculative decoding after
checking that benchmark accuracy is unchanged, because the engine documents them as exact.
Accuracy is an aggregate, so a feature can hold it while flipping which requests are answered
correctly. We serve one 4B instruction model in vLLM under nine single feature configurations and
five replicas of a reference configuration, grade \DNItems{} requests from five benchmarks in every
run, and compare each candidate with a restart floor (fresh process) and a scheduling floor (same
requests in a different order). With default kernels every configuration keeps accuracy within one
point, and the seven exact features flip the graded outcome of \DBsCorr\pp{} (95\pp{} CI \DBsCorrLo{}
to \DBsCorrHi) to \DNgramCorr\pp{} (\DNgramCorrLo{} to \DNgramCorrHi) of requests, none above the
scheduling floor of \DFloorSCorr\pp, while the FP8 KV cache flips \DFpkvCorr\pp{} (\DFpkvCorrLo{} to
\DFpkvCorrHi), as many as FP8 weights. With batch invariant kernels all seven exact features,
speculative decoding included, reproduce the reference text on every request, while the FP8 KV cache
still flips \BFpkvCorr\pp{} (\BFpkvCorrLo{} to \BFpkvCorrHi) of graded outcomes. A gate that compares
against a single replica pair rejected an unchanged configuration, so we recommend certifying a
serving change under batch invariant kernels, where the floor is zero, or against a scheduling floor
measured from at least three replicas.
\end{abstract}

\section{Introduction}
\label{sec:intro}

A production LLM service turns on serving features to cut latency and cost. vLLM, for example,
offers prefix caching, chunked prefill, CUDA graph capture, continuous batching with a configurable
number of concurrent sequences, speculative decoding with n-gram, EAGLE and draft model proposers,
and an FP8 KV cache \citep{kwon2023vllm,leviathan2022spec,li2025eagle3}. Most of these are described
as leaving the output distribution unchanged; speculative decoding in particular is lossless by
construction under exact arithmetic \citep{leviathan2022spec}. Teams therefore validate such a change
by rerunning a benchmark and checking that accuracy holds.

Floating point arithmetic in a batched engine is not exact. A request's logits depend on how many
other requests share its batch, on how its prompt is split into prefill chunks, and on which kernel
the engine picks for the shape at hand \citep{tml2025nondeterminism}. Under greedy decoding a change in
the last bit of a logit can switch one token, and one switched token can send a math derivation down a
different path. A feature can then leave accuracy at the same value while flipping which requests are
right. One GSM8K word problem from our runs shows it. All three restarts of the reference configuration
answer it correctly with \ExRefAns. With prefix caching on, the generated text matches the reference
for its first \ExSharedPct\pp{} of characters, then inserts one word into a restated sentence, and
the derivation that follows ends at \ExCandAns. Other items flip the other way, and accuracy under
prefix caching moves by \DPcDacc{} points.

Measuring this is harder than counting differences. With default kernels the same configuration
restarted in a fresh process already flips \DFloorRText\pp{} of generated texts and
\DFloorRCorr\pp{} of graded outcomes in our setup, and changing only the order in which the same
requests arrive raises both. A feature's flips mean something only above these floors, and the floors
must be estimated with intervals, from more than one pair of runs, on the same items.

We serve Qwen3 4B Instruct \citep{yang2025qwen3} in vLLM 0.33.0 on one GPU, grade \DNItems{} items
from GSM8K, MATH500, MMLU, HumanEval and MBPP
\citep{cobbe2021gsm8k,hendrycks2021math,hendrycks2020mmlu,chen2021humaneval,austin2021mbpp}, and run each
of nine single feature configurations once and the reference configuration five times (three restarts
and two reorderings). We repeat the design with vLLM's batch invariant kernels, which remove the
dependence of a request's arithmetic on its batch, as a control arm. The hypotheses, metrics and
decision rules were preregistered before any main run.

\begin{figure*}[t]
\centering
\includegraphics[width=\textwidth]{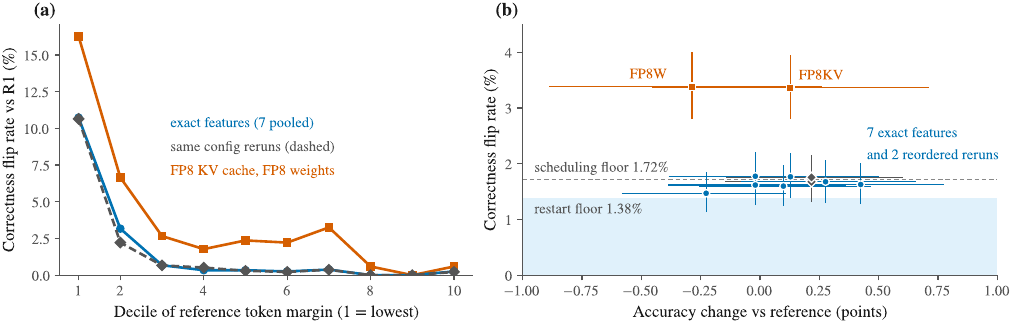}
\caption{Exact serving features flip graded answers as often as reordering the same requests does,
and the FP8 KV cache doubles the rate while accuracy holds. (a) Flip rate against R1 by decile of
R1's 10th percentile token margin (top 1 minus top 2 logprob), pooled per group. (b) Accuracy change
against flip rate per configuration, against R1 to R3; band: restart floor, dashed: scheduling floor,
bars: 95\pp{} cluster bootstrap intervals. Default kernels, \DNItems{} items per configuration.}
\label{fig:main}
\end{figure*}

\textbf{Contributions.}
\begin{itemize}
  \item \textbf{Exact features flip graded answers, but no more than reordering does.} With default
  kernels the seven exact features flip \DBsCorr\pp{} to \DNgramCorr\pp{} of graded outcomes while
  accuracy stays within one point; reordering the same requests flips \DSoneCorr\pp{} and
  \DStwoCorr\pp{}, and no feature exceeds the scheduling floor (\cref{fig:main}, \cref{sec:main}).
  \item \textbf{The FP8 KV cache is not near exact for graded answers.} It flips \DFpkvCorr\pp{}
  of outcomes, \DFpkvCorrExS{} points (one sided 95\pp{} lower bound \DFpkvCorrExSLo) above the
  scheduling floor and as many as FP8 weights, while its accuracy change of \DFpkvDacc{} points passes
  a one point equivalence test (\cref{sec:main}).
  \item \textbf{Under batch invariant kernels the exact features are exact.} Restarts and reorderings
  are bitwise identical, the seven exact features reproduce the reference text on all \BNItems{} items,
  and FP8W and FP8KV still flip \BFpwCorr\pp{} and \BFpkvCorr\pp{} of graded outcomes
  (\cref{fig:bi}, \cref{sec:bi}).
  \item \textbf{A flip gate needs a multi replica floor.} Holding out one restart replica against a
  floor from the other two produced a FAIL on an unchanged configuration, because single replica pairs
  range from \DPairMinCorr{} to \DPairMaxCorr{} flipped items (\cref{sec:gate}).
\end{itemize}

\begin{table}[t]
\caption{What a team checks before enabling a serving feature, and what we recommend.}
\label{tab:practice}
\centering\small
\begin{tabular}{@{}p{0.26\columnwidth}p{0.50\columnwidth}l@{}}
\toprule
Today & Recommendation & Section \\
\midrule
Accuracy unchanged & Paired accuracy equivalence test with a stated margin & \ref{sec:main} \\
Spot check outputs & Graded flip rate against a scheduling floor, not a restart floor & \ref{sec:main} \\
One rerun as control & At least three replicas, all pairs, cluster bootstrap & \ref{sec:gate} \\
\bottomrule
\end{tabular}
\end{table}

\section{Setup}
\label{sec:setup}

\textbf{Model, engine and hardware.} Qwen3 4B Instruct 2507 in bf16, served by vLLM 0.33.0 through
its offline \texttt{LLM.generate} interface in the official container, on one RTX 5070 Ti (16 GB).
Decoding is greedy (temperature 0, seed 0). All requests of a task are submitted together, so the
engine forms batches by continuous batching as it would under load. Each configuration runs in a
fresh process.

\textbf{Items and grading.} \DNItems{} items: all 1319 GSM8K test items (four shot), all 500
MATH500 items (boxed answer, graded with math-verify), 1140 MMLU test items (20 per subject, five shot,
answer generated in free form with a final answer line), all 164 HumanEval items and all 257 sanitized
MBPP items (graded by executing the official tests in a container without network). Generation caps are
768, 4096, 1024, 1024 and 1024 tokens; a truncated output is graded as produced. The decision unit is
the item, and its decision is the graded outcome (right or wrong).

\textbf{Configurations.} The reference configuration (REF) has prefix caching off, a prefill budget of
8192 tokens, at most 128 concurrent sequences, CUDA graphs on, a bf16 KV cache and no speculation.
Each candidate changes one factor: prefix caching on (PC); a prefill budget of 512 tokens so that every
GSM8K prompt and most MMLU prompts are prefilled in chunks (CP); at most 16 concurrent sequences (BS);
CUDA graphs off (EAGER); speculative decoding with an n-gram proposer (NGRAM, four tokens), an EAGLE-3
head (EAGLE3, three tokens) or Qwen3 0.6B as draft model (DRAFT, four tokens); an FP8 KV cache (FP8KV);
and dynamic FP8 weights (FP8W), a positive control that is known to be lossy. We call PC, CP, BS, EAGER,
NGRAM, EAGLE3 and DRAFT the exact features: under exact arithmetic none of them changes the greedy
output. REF runs five times: R1, R2 and R3 in fresh processes with the canonical request order, and S1
and S2 with the request order shuffled by seeds 1 and 2, which changes batch composition but nothing
else. The batch invariant arm repeats the design with \texttt{VLLM\_BATCH\_INVARIANT=1}.

\textbf{Metrics.} For a candidate C and reference replicas $R_1,\dots,R_k$, the per item flip probability
is the share of the $k$ pairs (C, $R_j$) in which the item's graded outcome differs, which we call a
flip; the correctness flip rate (flip rate for short) is its mean over items. The restart floor is the same quantity over all pairs of restart replicas,
and the scheduling floor over all pairs (S, R). The excess flip rate is the per item flip probability
minus the per item floor, averaged over items, so both rates are paired on the same items. Accuracy
change is the paired difference of candidate and mean reference accuracy. All intervals are cluster
bootstrap intervals (2000 resamples, seed 0; cluster is the MMLU subject, otherwise the item,
\DNClusters{} clusters). Secondary metrics are the answer flip rate (extracted answer differs) and the
text flip rate (any generated token differs).

\textbf{Preregistered claims and rules.} C1: an exact feature ``flips'' if the one sided 95\pp{} lower
bound of its excess over the restart floor is above zero, and is accuracy equivalent if the 90\pp{}
interval of its accuracy change lies within one point; C1 holds if at least three of the seven exact
features do both. C2: under batch invariant kernels PC, CP, BS and EAGER produce zero text flips, the
speculative features produce zero correctness flips, and both FP8 configurations produce some. C3: the
gate (\cref{sec:gate}) gives no FAIL on any null comparison and FAIL on FP8W, in each arm. Everything
else is labelled exploratory.

\section{The instrument reads zero where it should}
\label{sec:kill}

\textbf{Grading is deterministic (K1).} Rescoring the outputs of \KoneRescoreRuns{} runs gives
\KoneRescoreDiff{} differing rows out of \KoneRescoreRows; re executing the \KoneExecProgs{} generated
programs of R1 in a fresh sandbox gives \KoneExecDiff{} differing pass flags.

\textbf{Every feature is engaged (K4).} Engine counters, read before and after each task, show
\DKPcHits{} prefix cache hit tokens for PC, \DKCpOver{} prompts above the 512 token prefill budget for CP,
and draft acceptance rates of \DKNgramAcc\pp{} (NGRAM), \DKEagleAcc\pp{} (EAGLE3) and \DKDraftAcc\pp{}
(DRAFT). No feature is excluded.

\textbf{The default kernel floor is under the flag line (K3).} Restarting REF flips
\DFloorRCorr\pp{} of graded outcomes (one sided 95\pp{} upper bound \DFloorRCorrUp\pp{}), below the 2\pp{}
line at which the preregistration flags the default arm as unstable. It flips \DFloorRAns\pp{} of
extracted answers and \DFloorRText\pp{} of generated texts. Reordering raises the three rates to
\DFloorSCorr\pp, \DFloorSAns\pp{} and \DFloorSText\pp.

\textbf{Batch invariant restarts are bitwise identical (K2).} With \texttt{VLLM\_BATCH\_INVARIANT=1},
the three restart pairs of REF differ on \BPairMaxText{} generated texts, so the batch invariant restart
floor is zero for texts, answers and grades (one sided 95\pp{} upper bound \BFloorRTextUp\pp{} of items).
The two reordered replicas also match every restart replica token for token (scheduling floor
\BFloorSText\pp). K2 passes, and C2 is scored as preregistered.

\section{Exact features flip answers at the scheduling floor}
\label{sec:main}

\begin{table*}[t]
\caption{Every configuration keeps accuracy equivalent, the exact features flip graded outcomes as
often as a reordered rerun, and the FP8 configurations double the rate. Default kernels, \DNItems{}
items, flip rate against R1 to R3 with 95\pp{} cluster bootstrap interval. Excess over the restart floor
(\DFloorRCorr\pp): one sided 95\pp{} lower bound; over the scheduling floor (\DFloorSCorr\pp): estimate,
one sided 95\pp{} bounds. $\Delta$acc: 90\pp{} interval, margin one point. Text: text flip rate. Gate:
\cref{sec:gate}, 0.5 point tolerance (INCONC.\ is inconclusive).}
\label{tab:default}
\centering\small\setlength{\tabcolsep}{4.5pt}
\begin{tabular}{@{}lrlrrlrlrl@{}}
\toprule
& \multicolumn{2}{c}{Flip rate (\%)} & \multicolumn{3}{c}{Excess (points)} & \multicolumn{2}{c}{$\Delta$acc (points)} & Text & \\
\cmidrule(lr){2-3}\cmidrule(lr){4-6}\cmidrule(lr){7-8}
Configuration & est. & 95\% CI & restart lower & sched.\ est. & sched.\ bounds & est. & 90\% CI & (\%) & Gate \\
\midrule
\input{tab_default.tex}
\end{tabular}
\end{table*}

\textbf{Accuracy holds everywhere.} Reference accuracy is \DAccRef\pp{} (95\pp{} CI \DAccRefLo{} to
\DAccRefHi). Every configuration, the FP8
positive control included, has its 90\pp{} accuracy interval inside the one point margin
(\cref{tab:default}). The largest change is EAGLE3 at \DEagleDacc{} points (90\pp{} CI \DEagleDaccLo{} to
\DEagleDaccHi), which is inside the margin although its interval excludes zero. An accuracy check alone
would pass all eleven candidates.

\textbf{Exact features flip graded outcomes.} The seven exact features flip the graded outcome of
\DBsCorr\pp{} (BS) to \DNgramCorr\pp{} (NGRAM) of items and the generated text of \DBsText\pp{}
(95\pp{} CI \DBsTextLo{} to \DBsTextHi) to \DEagerText\pp{} (\DEagerTextLo{} to \DEagerTextHi). Three of them, EAGER, NGRAM and DRAFT, have a one sided lower bound on their excess over
the restart floor above zero (\DEagerCorrExRLo, \DNgramCorrExRLo{} and \DDraftCorrExRLo{} points) and are
accuracy equivalent. By the preregistered rule C1 holds: three of seven, exactly the threshold.

\textbf{The scheduling floor explains the excess.} The two reordered replicas of REF also pass the C1
rule (lower bounds \DSoneCorrExRLo{} and \DStwoCorrExRLo), although nothing in their configuration
changed. Against the scheduling floor, no exact feature has a positive lower bound: the largest
estimates are \DNgramCorrExS{} (NGRAM) and \DEagerCorrExS{} points (EAGER), with upper bounds of
\DNgramCorrExSUp{} and \DEagerCorrExSUp. C1 therefore holds as preregistered, but the excess it detects
is of the size that a change in request order produces on its own. For the exact features with default
kernels we read this as: they flip answers, and so does production traffic.

\textbf{The FP8 KV cache behaves like FP8 weights.} FP8KV flips \DFpkvCorr\pp{} of graded outcomes
(95\pp{} CI \DFpkvCorrLo{} to \DFpkvCorrHi) and FP8W \DFpwCorr\pp{} (\DFpwCorrLo{} to \DFpwCorrHi), each
about \DFpkvCorrExS{} points above the scheduling floor, and each flips more than half of all generated
texts (\DFpkvText\pp{} and \DFpwText\pp). The direction is balanced: against R1, FP8KV makes
\DFpkvBetter{} items right and \DFpkvWorse{} wrong (exact McNemar $p = \DFpkvMcN$), which is why accuracy
holds. Per task, the FP8 configurations raise the MMLU flip rate from at most \DPcCorrMmlu\pp{} under the
exact features to \DFpkvCorrMmlu\pp{} and \DFpwCorrMmlu\pp, so short answer tasks are where the KV cache
precision shows most clearly (\cref{app:pertask}).

\textbf{Long math derivations carry most flips.} MATH500 has the highest flip rate for every
configuration, \DBsCorrMath\pp{} to \DDraftCorrMath\pp{} for the exact features, because a 4096 token
derivation gives one switched token many chances to flip the result.

\subsection{Seed noise and nondeterminism do not explain the FP8 KV cache}

Greedy decoding with a fixed seed removes sampling noise, and K1 removes grading noise. What remains is
numerical: batch composition, kernel selection and preemption. Both floors include all three, since
every REF replica sees about \DKRonePreempt{} preemptions, as most configurations do (the BS configuration,
with at most 16 sequences, sees \DKBsPreempt). FP8KV's excess over the scheduling floor has a one sided
lower bound of \DFpkvCorrExSLo{} points, so the cheap explanations do not account for it. The batch
invariant arm removes them outright: its floor is exactly zero, and FP8KV still flips \BFpkvCorr\pp{}
of graded outcomes there (\cref{sec:bi}).

\section{Batch invariant kernels remove every exact feature flip}
\label{sec:bi}

\begin{figure*}[t]
\centering
\includegraphics[width=\textwidth]{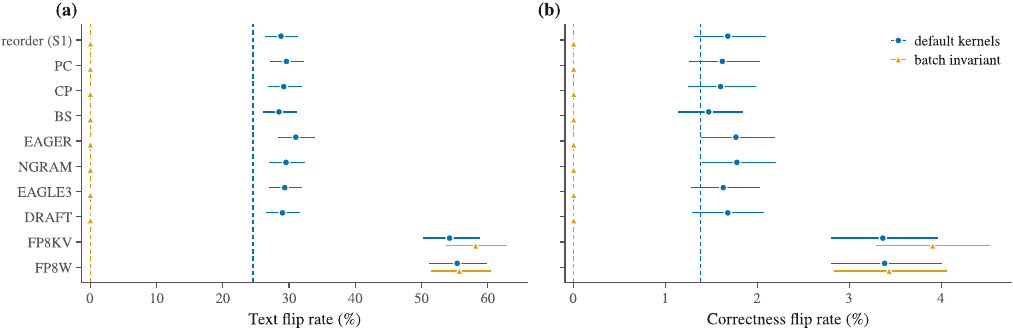}
\caption{Batch invariant kernels remove every flip of the exact features and leave the FP8 flips in
place. (a) Text flip rate and (b) correctness flip rate of each configuration against R1, R2 and R3,
under default kernels (circles) and batch invariant kernels (triangles); \DNItems{} items per
configuration, bars are 95\pp{} cluster bootstrap intervals, dashed lines the restart floor of each arm.
Under batch invariant kernels the reordered replica and all seven exact features reproduce the reference
text on every item.}
\label{fig:bi}
\end{figure*}

With \texttt{VLLM\_BATCH\_INVARIANT=1}, vLLM computes each request with kernels whose reduction order
does not depend on the batch the request shares \citep{tml2025nondeterminism}. If the default arm flips
of the exact features come from batch dependent arithmetic, they should vanish under these kernels;
flips caused by lower precision should not. We ran all fourteen configurations again in this mode
(\cref{fig:bi}, \cref{tab:bi}). Every feature stays engaged (K4): \BKPcHits{} prefix cache hit tokens for
PC, \BKCpOver{} prompts above the prefill budget for CP, and draft acceptance rates of \BKNgramAcc\pp,
\BKEagleAcc\pp{} and \BKDraftAcc\pp{} for NGRAM, EAGLE3 and DRAFT.

\textbf{The exact features are exact in arithmetic.} PC, CP, BS and EAGER produce zero text flips against
all three restart replicas, as C2 predicts. For NGRAM, EAGLE3 and DRAFT we predicted only zero
correctness flips; they too reproduce the reference text token for token on all \BNItems{} items, which
is what speculative decoding promises under exact arithmetic \citep{leviathan2022spec}. With zero
flips in \BNItems{} items, the one sided 95\pp{} upper bound on each feature's text flip rate is
\BFloorRTextUp\pp{} of items.

\textbf{FP8 still flips.} FP8KV flips \BFpkvCorr\pp{} of graded outcomes (95\pp{} CI \BFpkvCorrLo{} to
\BFpkvCorrHi, \BFpkvCorrItems{} items) and FP8W \BFpwCorr\pp{} (\BFpwCorrLo{} to \BFpwCorrHi,
\BFpwCorrItems{} items), and each flips more than half of all texts (\BFpkvText\pp{} and
\BFpwText\pp). With a zero floor these rates belong to the precision change alone, and they are close
to the default arm rates (\DFpkvCorr\pp{} and \DFpwCorr\pp). C2 holds for all nine features. One
result differs from the default arm: the FP8KV accuracy change of \BFpkvDacc{} points has a 90\pp{}
interval from \BFpkvDaccLo{} to \BFpkvDaccHi, which crosses the one point margin, so its accuracy
equivalence is inconclusive in this arm, while FP8W stays equivalent (\BFpwDacc{} points, \BFpwDaccLo{}
to \BFpwDaccHi).

\textbf{Where the default arm flips come from.} Under default kernels an exact feature flips about as
many outcomes as reordering the same requests; under batch invariant kernels neither switches a single
token. The exact features therefore flip answers only through the batch dependence of the
arithmetic, the same channel that request order uses, whereas FP8 flips them through precision in
either arm.

\textbf{What the control costs (exploratory, indicative).} REF generates about \BiCostPct\pp{} slower
with batch invariant kernels than with default kernels (mean of three restarts per arm, shared GPU).
Within the batch invariant arm the speedups keep their default arm pattern: FP8W \BFpwSpeed{}, FP8KV
\BFpkvSpeed{}, PC \BPcSpeed{} and EAGLE3 \BEagleSpeed{} times as fast as REF. An audit run in this mode
needs no floor estimate: one candidate run against one reference run attributes every flip to the
change.

\section{A flip gate needs a multi replica floor}
\label{sec:gate}

\begin{table}[t]
\caption{A floor from one replica pair fails an unchanged configuration; batch invariant kernels pass
every null comparison. Gate verdict and excess flip rate over the floor (estimate, one sided 95\pp{}
upper bound, points) for each null comparison; R$k$ is held out against the other two restarts, S1
and S2 are compared with all three. Tolerance 0.5 points, \DNItems{} items; INCONC.\ is inconclusive.}
\label{tab:nulls}
\centering\small
\begin{tabular}{@{}lrrlrl@{}}
\toprule
& \multicolumn{3}{c}{Default kernels} & \multicolumn{2}{c}{Batch invariant} \\
\cmidrule(lr){2-4}\cmidrule(lr){5-6}
Held out & est. & upper & Gate & est. & Gate \\
\midrule
R1 & \DNullRoneEx & \DNullRoneUp & \DNullRone & \BNullRoneEx & \BNullRone \\
R2 & \DNullRtwoEx & \DNullRtwoUp & \DNullRtwo & \BNullRtwoEx & \BNullRtwo \\
R3 & \DNullRthreeEx & \DNullRthreeUp & \DNullRthree & \BNullRthreeEx & \BNullRthree \\
S1 & \DNullSoneEx & \DNullSoneUp & \DNullSoneShort & \BNullSoneEx & \BNullSone \\
S2 & \DNullStwoEx & \DNullStwoUp & \DNullStwoShort & \BNullStwoEx & \BNullStwo \\
\bottomrule
\end{tabular}
\end{table}

We ship the decision rule as a command line gate. Given reference replicas and candidate runs on a
fixed request set, it reports the paired accuracy change with its 90\pp{} interval (PASS when inside a
margin, here one point) and the excess flip rate over the floor of all reference replica pairs (PASS
when the one sided 95\pp{} upper bound is at most a tolerance, here 0.5 points; FAIL when the lower bound
is above it; otherwise INCONCLUSIVE). The overall verdict is FAIL if either test fails.

\textbf{The gate catches both FP8 configurations.} FP8W and FP8KV get FAIL on the flip test while
passing the accuracy test; CP and BS get PASS; the other exact features are INCONCLUSIVE at this sample
size (\cref{tab:default}).

\textbf{A held out restart replica failed.} For the preregistered null comparisons each restart replica
is held out as a candidate against the other two (\cref{tab:nulls}). R1 and R3 get PASS; R2 gets FAIL, with an excess upper
bound of \DNullRtwoUp{} points. The cause is the floor: with two remaining replicas the floor comes from a
single pair, and the three restart pairs differ on \DPairMinCorr{} (\DPairMinName) to \DPairMaxCorr{}
(\DPairMaxName) items, that is \DPairMinPct\pp{} to \DPairMaxPct\pp{} of items. Holding out R2 leaves the
quietest pair as the floor. C3 is therefore refuted in the default arm. The reordered replicas S1 and S2,
compared against all three restart replicas, get INCONCLUSIVE. The lesson for practice is concrete: a
floor from one replica pair is itself a noisy estimate, so a gate needs at least three replicas, and in
the default arm it should include reordered replicas, since production traffic reorders requests anyway.

\textbf{With batch invariant kernels the gate separates cleanly.} All five null comparisons (each
restart replica held out, and S1 and S2) get PASS, every exact feature gets PASS, and FP8W and FP8KV get
FAIL on the flip test. C3's prediction holds in this arm. Because the preregistered claim requires it in
each arm, C3 as a whole stays refuted by the default arm result.

\section{Which requests flip (exploratory)}
\label{sec:margin}

Flips concentrate on items the reference answers with a small token margin
(\cref{fig:main}a). The 10th percentile of the top 1 minus top 2 logprob along R1's output predicts
whether an item flips under an exact feature with ROC AUC \DExactAucQ{} (95\pp{} CI \DExactAucQLo{} to
\DExactAucQHi), and the lowest decile of that margin holds \DExactLowShare\pp{} (\DExactLowShareLo{} to
\DExactLowShareHi) of exact feature flips.
The same margin predicts restart and reorder flips equally well (AUC \DFloorAucQ), and FP8 flips less well
(AUC \DLossyAucQ), because FP8 also flips items with a moderate margin. Two caveats keep this exploratory.
The minimum margin is exactly zero for \DMarginZero\pp{} of items, because logprobs are returned at bf16
resolution, so it cannot rank the tail (AUC \DExactAucMin). And output length alone predicts exact feature
flips with AUC \DExactAucLen{} (\DExactAucLenLo{} to \DExactAucLenHi), since the margin and log length
correlate at \DCorrLenMargin. A margin based triage adds little to ``long outputs flip''.

\textbf{Speed beside the flips (exploratory, indicative).} On this shared GPU, relative to REF, FP8W
and FP8KV generate \DFpwSpeed{} and \DFpkvSpeed{} times as fast, PC \DPcSpeed{} times and EAGLE3
\DEagleSpeed{} times; NGRAM (\DNgramSpeed), BS (\DBsSpeed), EAGER (\DEagerSpeed) and DRAFT (\DDraftSpeed)
are not faster at 128 concurrent sequences. The two configurations with the largest speedup are the two
that double the flip rate.

\section{Related work}
\label{sec:related}

\textbf{Noise floors and quantization flips.} \citet{blagojevich2026floor} is closest in method: it
measures restart, engine, batch and kernel drift and states 8 bit and 4 bit weight quantization effects
as excess over it. Its treatment is weight quantization; we toggle the serving features themselves,
separate a restart floor from a scheduling floor, and add a batch invariant control arm.
\citet{wu2026lowbit} predict which decisions low bit weight quantization breaks from the decision
margin; our exploratory margin result agrees in direction but finds output length as predictive for
generative tasks. \citet{chen2026alibi} localizes divergence from one INT8 kernel swap.

\textbf{Single serving features.} \citet{patodiya2026cache} toggles prefix caching in multi turn agentic
episodes at batch size one and finds divergence that grows with quantization while accuracy holds;
\citet{zhang2025forensics} shows that batch speculative decoding implementations can corrupt outputs and
that vLLM speculative output matches its non speculative output on 65.6\pp{} to 96.9\pp{} of prompts by
exact text match; \citet{chodavarapu2026illusion} compares KV cache on and off in FP16.
\citet{xu2026alignment} show that KV cache quantization, including FP8 KV caches served in vLLM, can
remove refusals while perplexity barely moves, the same pattern of a passing aggregate metric over
changed individual outputs that we measure for graded answers. We measure graded
correctness rather than text equality, for seven features in one engine, against two floors.

\textbf{Drift across stacks.} DriftBench \citep{vitale2026driftbench} measures flip rates across
hardware, frameworks and precisions and blocks an upgrade with a regression predictor;
\citet{pape2026silent} show that the inference backend moves benchmark accuracy. Both vary the stack
broadly; we hold the engine, model and GPU fixed and vary one documented feature at a time.
\citet{tml2025nondeterminism} explains batch dependence and provides the batch invariant kernels that
vLLM exposes, which we use as a control rather than as the object of study.

\section{Limitations}
\label{sec:limits}

\begin{enumerate}
  \item One model (a 4B instruction model), one engine version and one consumer GPU. Larger models
  have larger margins on easy items and longer outputs on hard ones; we did not test which dominates.
  \item One candidate run per feature. The excess uses all reference replicas, but the candidate's own
  restart variability is not averaged out, which widens rather than biases the interval.
  \item Three restart replicas. The gate result in \cref{sec:gate} shows that the floor estimate is
  itself variable; more replicas would narrow every excess interval.
  \item Request order is the only scheduling change we control. Real traffic also changes arrival
  times and concurrent request lengths, which we expect to raise the scheduling floor further.
  \item The GPU was shared with other jobs during the runs, so the speed numbers are indicative only;
  flip rates do not depend on timing beyond what the floors already include.
  \item The batch invariant result holds for the kernels of vLLM 0.33.0. A feature that falls back to a
  batch dependent kernel in another version or on another GPU would not be exact there.
  \item Five benchmarks with exact grading. Tasks graded by a judge or by human preference may react
  differently.
\end{enumerate}

We do not claim that these features make the model less accurate: every default arm configuration is
accuracy equivalent. The claim is about which requests are answered correctly.

\section{Conclusion}

For one model in vLLM with default kernels, the serving features documented as exact flip the graded
answer of \DBsCorr\pp{} to \DNgramCorr\pp{} of requests, which matches what reordering the same
requests does, and the FP8 KV cache flips \DFpkvCorr\pp, as many as FP8 weights, with accuracy
equivalent in every case. A team that enables a serving feature should compare it with a scheduling
floor measured from at least three replicas, report the paired flip rate with a cluster bootstrap
interval next to the accuracy change, and treat the FP8 KV cache as a lossy change. With batch
invariant kernels the same features switch no generated token, so running the audit in that mode
turns a noisy floor estimate into a zero floor, and any flip that remains belongs to the change.

\section*{Impact Statement}
Serving features are switched on in production to cut latency and cost, usually after an accuracy
check. When they flip which requests are answered correctly, individual users get different answers
from a model whose reported accuracy did not move, and nobody chose that change on purpose. The gate
and the batch invariant audit described here let a team see and bound those flips before a rollout.
The same measurement could be used to argue that a cheaper configuration is safe on accuracy alone;
reporting the flip rate next to the accuracy change makes that choice visible. We use public
benchmarks and open weight models and collect no human data.

\section*{AI Disclosure Statement}
AI assistants were used for code and experiment orchestration, and for drafting the text. Every number
in the paper is produced by a script from the result files and enters the text as a generated macro,
so no number is typed by hand, and every reference was checked against its arXiv or publisher record.
The authors reviewed all code, results and text and take full responsibility for them.

\bibliography{refs}
\bibliographystyle{icml2027}

\newpage
\appendix
\onecolumn

\section{Per task correctness flip rates}
\label{app:pertask}

\begin{table}[h]
\caption{MATH500 carries the most flips for every configuration; the FP8 configurations raise the
MMLU and code flip rates most. Correctness flip rate (\%) against R1, R2 and R3 per task, default
kernels; items per task: GSM8K 1319, MATH500 500, MMLU 1140, HumanEval 164, MBPP 257.}
\label{tab:pertask}
\centering\small
\begin{tabular}{@{}lrrrrr@{}}
\toprule
Configuration & GSM8K & MATH500 & MMLU & HumanEval & MBPP \\
\midrule
Reordered S1 & \DSoneCorrGsm & \DSoneCorrMath & \DSoneCorrMmlu & \DSoneCorrHe & \DSoneCorrMbpp \\
Reordered S2 & \DStwoCorrGsm & \DStwoCorrMath & \DStwoCorrMmlu & \DStwoCorrHe & \DStwoCorrMbpp \\
\midrule
Prefix caching & \DPcCorrGsm & \DPcCorrMath & \DPcCorrMmlu & \DPcCorrHe & \DPcCorrMbpp \\
Chunked prefill 512 & \DCpCorrGsm & \DCpCorrMath & \DCpCorrMmlu & \DCpCorrHe & \DCpCorrMbpp \\
Max 16 sequences & \DBsCorrGsm & \DBsCorrMath & \DBsCorrMmlu & \DBsCorrHe & \DBsCorrMbpp \\
No CUDA graphs & \DEagerCorrGsm & \DEagerCorrMath & \DEagerCorrMmlu & \DEagerCorrHe & \DEagerCorrMbpp \\
Spec.\ n-gram & \DNgramCorrGsm & \DNgramCorrMath & \DNgramCorrMmlu & \DNgramCorrHe & \DNgramCorrMbpp \\
Spec.\ EAGLE-3 & \DEagleCorrGsm & \DEagleCorrMath & \DEagleCorrMmlu & \DEagleCorrHe & \DEagleCorrMbpp \\
Spec.\ draft model & \DDraftCorrGsm & \DDraftCorrMath & \DDraftCorrMmlu & \DDraftCorrHe & \DDraftCorrMbpp \\
\midrule
FP8 KV cache & \DFpkvCorrGsm & \DFpkvCorrMath & \DFpkvCorrMmlu & \DFpkvCorrHe & \DFpkvCorrMbpp \\
FP8 weights & \DFpwCorrGsm & \DFpwCorrMath & \DFpwCorrMmlu & \DFpwCorrHe & \DFpwCorrMbpp \\
\bottomrule
\end{tabular}
\end{table}

\section{Batch invariant arm}
\label{app:bi}

\begin{table}[h]
\caption{With batch invariant kernels every same configuration rerun and every exact feature reproduces
the reference text on all items, and the FP8 configurations flip graded outcomes at the default arm
rate. Columns as in \cref{tab:default}; restart and scheduling floors are both \BFloorRCorr\pp{},
\BNItems{} items.}
\label{tab:bi}
\centering\small\setlength{\tabcolsep}{4.5pt}
\begin{tabular}{@{}lrlrrlrlrl@{}}
\toprule
& \multicolumn{2}{c}{Flip rate (\%)} & \multicolumn{3}{c}{Excess (points)} & \multicolumn{2}{c}{$\Delta$acc (points)} & Text & \\
\cmidrule(lr){2-3}\cmidrule(lr){4-6}\cmidrule(lr){7-8}
Configuration & est. & 95\% CI & restart lower & sched.\ est. & sched.\ bounds & est. & 90\% CI & (\%) & Gate \\
\midrule
\input{tab_bi.tex}
\end{tabular}
\end{table}

\section{Preregistration and amendment}
\label{app:prereg}

The preregistration (thesis, claims C1 to C3, metrics, thresholds, kill checks K1 to K4 and the
exploratory list) was committed before any main run, at commit
\texttt{c00e671c0298b57e062646fe951afdd710fb5175}. One dated amendment followed after the default arm
was scored and touches exploratory analyses only: the margin analysis uses the 10th percentile margin
for the decile summaries and Figure 1(a), because the minimum margin is exactly zero for many items at
bf16 logprob resolution, and output length is added as a comparison predictor. No confirmatory claim,
metric, threshold or decision rule changed. The preregistration with its amendment, the code, the per item outputs of every run and the
scripts that produce every number, table and figure will be released with the public code release and are available from the author on request until then.

\end{document}

%% file: numbers.tex
\newcommand{\BEagleSpeed}{1.09}

\newcommand{\BFloorRCorr}{0.00}

\newcommand{\BFloorRTextUp}{0.09}

\newcommand{\BFloorSText}{0.00}

\newcommand{\BFpkvCorr}{3.91}

\newcommand{\BFpkvCorrHi}{4.53}
\newcommand{\BFpkvCorrItems}{132}
\newcommand{\BFpkvCorrLo}{3.28}

\newcommand{\BFpkvDacc}{\ensuremath{-}0.47}
\newcommand{\BFpkvDaccHi}{+0.11}
\newcommand{\BFpkvDaccLo}{\ensuremath{-}1.02}

\newcommand{\BFpkvSpeed}{1.49}
\newcommand{\BFpkvText}{58.17}

\newcommand{\BFpwCorr}{3.43}

\newcommand{\BFpwCorrHi}{4.06}
\newcommand{\BFpwCorrItems}{116}
\newcommand{\BFpwCorrLo}{2.83}

\newcommand{\BFpwDacc}{\ensuremath{-}0.18}
\newcommand{\BFpwDaccHi}{+0.33}
\newcommand{\BFpwDaccLo}{\ensuremath{-}0.70}

\newcommand{\BFpwSpeed}{1.90}
\newcommand{\BFpwText}{55.71}

\newcommand{\BKCpOver}{1{,}982}
\newcommand{\BKDraftAcc}{65.7}
\newcommand{\BKEagleAcc}{51.1}
\newcommand{\BKNgramAcc}{13.6}
\newcommand{\BKPcHits}{1{,}424{,}800}

\newcommand{\BNItems}{3{,}380}

\newcommand{\BNullRone}{PASS}
\newcommand{\BNullRoneEx}{+0.00}

\newcommand{\BNullRthree}{PASS}
\newcommand{\BNullRthreeEx}{+0.00}

\newcommand{\BNullRtwo}{PASS}
\newcommand{\BNullRtwoEx}{+0.00}

\newcommand{\BNullSone}{PASS}
\newcommand{\BNullSoneEx}{+0.00}

\newcommand{\BNullStwo}{PASS}
\newcommand{\BNullStwoEx}{+0.00}

\newcommand{\BPairMaxText}{0}

\newcommand{\BPcSpeed}{1.23}

\newcommand{\BiCostPct}{4}
\newcommand{\DAccRef}{85.94}
\newcommand{\DAccRefHi}{87.53}
\newcommand{\DAccRefLo}{84.26}

\newcommand{\DBsCorr}{1.47}

\newcommand{\DBsCorrGsm}{1.54}
\newcommand{\DBsCorrHe}{0.61}
\newcommand{\DBsCorrHi}{1.85}

\newcommand{\DBsCorrLo}{1.13}
\newcommand{\DBsCorrMath}{4.53}
\newcommand{\DBsCorrMbpp}{1.17}
\newcommand{\DBsCorrMmlu}{0.23}

\newcommand{\DBsSpeed}{0.84}
\newcommand{\DBsText}{28.47}

\newcommand{\DBsTextHi}{31.16}

\newcommand{\DBsTextLo}{25.99}

\newcommand{\DCorrLenMargin}{\ensuremath{-}0.76}

\newcommand{\DCpCorrGsm}{1.64}
\newcommand{\DCpCorrHe}{0.41}

\newcommand{\DCpCorrMath}{5.13}
\newcommand{\DCpCorrMbpp}{1.30}
\newcommand{\DCpCorrMmlu}{0.23}

\newcommand{\DDraftCorrExRLo}{+0.05}

\newcommand{\DDraftCorrGsm}{1.42}
\newcommand{\DDraftCorrHe}{0.61}

\newcommand{\DDraftCorrMath}{6.53}
\newcommand{\DDraftCorrMbpp}{1.04}
\newcommand{\DDraftCorrMmlu}{0.15}

\newcommand{\DDraftSpeed}{0.50}

\newcommand{\DEagerCorrExRLo}{+0.12}

\newcommand{\DEagerCorrExS}{+0.05}

\newcommand{\DEagerCorrExSUp}{+0.32}
\newcommand{\DEagerCorrGsm}{1.79}
\newcommand{\DEagerCorrHe}{0.61}

\newcommand{\DEagerCorrMath}{5.33}
\newcommand{\DEagerCorrMbpp}{1.82}
\newcommand{\DEagerCorrMmlu}{0.32}

\newcommand{\DEagerSpeed}{0.77}
\newcommand{\DEagerText}{31.02}

\newcommand{\DEagerTextHi}{33.87}

\newcommand{\DEagerTextLo}{28.37}

\newcommand{\DEagleCorrGsm}{1.47}
\newcommand{\DEagleCorrHe}{1.02}

\newcommand{\DEagleCorrMath}{5.53}
\newcommand{\DEagleCorrMbpp}{1.04}
\newcommand{\DEagleCorrMmlu}{0.32}
\newcommand{\DEagleDacc}{+0.42}
\newcommand{\DEagleDaccHi}{+0.73}
\newcommand{\DEagleDaccLo}{+0.12}

\newcommand{\DEagleSpeed}{1.16}

\newcommand{\DExactAucLen}{0.88}
\newcommand{\DExactAucLenHi}{0.91}
\newcommand{\DExactAucLenLo}{0.84}
\newcommand{\DExactAucMin}{0.79}

\newcommand{\DExactAucQ}{0.89}
\newcommand{\DExactAucQHi}{0.92}
\newcommand{\DExactAucQLo}{0.85}

\newcommand{\DExactLowShare}{70}
\newcommand{\DExactLowShareHi}{79}
\newcommand{\DExactLowShareLo}{60}

\newcommand{\DFloorAucQ}{0.89}

\newcommand{\DFloorRAns}{3.01}

\newcommand{\DFloorRCorr}{1.38}

\newcommand{\DFloorRCorrUp}{1.69}
\newcommand{\DFloorRText}{24.56}

\newcommand{\DFloorSAns}{3.57}

\newcommand{\DFloorSCorr}{1.72}

\newcommand{\DFloorSText}{29.00}

\newcommand{\DFpkvBetter}{63}
\newcommand{\DFpkvCorr}{3.36}

\newcommand{\DFpkvCorrExS}{+1.65}
\newcommand{\DFpkvCorrExSLo}{+1.19}

\newcommand{\DFpkvCorrGsm}{2.48}
\newcommand{\DFpkvCorrHe}{4.88}
\newcommand{\DFpkvCorrHi}{3.96}

\newcommand{\DFpkvCorrLo}{2.80}
\newcommand{\DFpkvCorrMath}{6.60}
\newcommand{\DFpkvCorrMbpp}{3.89}
\newcommand{\DFpkvCorrMmlu}{2.63}
\newcommand{\DFpkvDacc}{+0.13}

\newcommand{\DFpkvMcN}{0.93}
\newcommand{\DFpkvSpeed}{1.66}
\newcommand{\DFpkvText}{54.23}

\newcommand{\DFpkvWorse}{61}

\newcommand{\DFpwCorr}{3.38}

\newcommand{\DFpwCorrGsm}{2.53}
\newcommand{\DFpwCorrHe}{2.64}
\newcommand{\DFpwCorrHi}{4.01}

\newcommand{\DFpwCorrLo}{2.80}
\newcommand{\DFpwCorrMath}{7.13}
\newcommand{\DFpwCorrMbpp}{4.54}
\newcommand{\DFpwCorrMmlu}{2.57}

\newcommand{\DFpwSpeed}{1.77}
\newcommand{\DFpwText}{55.37}

\newcommand{\DKBsPreempt}{34}
\newcommand{\DKCpOver}{1{,}982}
\newcommand{\DKDraftAcc}{65.6}
\newcommand{\DKEagleAcc}{50.8}
\newcommand{\DKNgramAcc}{13.7}
\newcommand{\DKPcHits}{1{,}424{,}672}

\newcommand{\DKRonePreempt}{1{,}947}

\newcommand{\DLossyAucQ}{0.78}

\newcommand{\DMarginZero}{38}
\newcommand{\DNClusters}{2297}
\newcommand{\DNItems}{3{,}380}

\newcommand{\DNgramCorr}{1.78}

\newcommand{\DNgramCorrExRLo}{+0.14}

\newcommand{\DNgramCorrExS}{+0.06}

\newcommand{\DNgramCorrExSUp}{+0.33}
\newcommand{\DNgramCorrGsm}{1.69}
\newcommand{\DNgramCorrHe}{0.61}
\newcommand{\DNgramCorrHi}{2.20}

\newcommand{\DNgramCorrLo}{1.39}
\newcommand{\DNgramCorrMath}{6.00}
\newcommand{\DNgramCorrMbpp}{1.17}
\newcommand{\DNgramCorrMmlu}{0.32}

\newcommand{\DNgramSpeed}{0.97}

\newcommand{\DNullRone}{PASS}
\newcommand{\DNullRoneEx}{\ensuremath{-}0.28}

\newcommand{\DNullRoneUp}{\ensuremath{-}0.03}
\newcommand{\DNullRthree}{PASS}
\newcommand{\DNullRthreeEx}{\ensuremath{-}0.68}

\newcommand{\DNullRthreeUp}{\ensuremath{-}0.45}
\newcommand{\DNullRtwo}{FAIL}
\newcommand{\DNullRtwoEx}{+0.96}

\newcommand{\DNullRtwoUp}{+1.34}

\newcommand{\DNullSoneEx}{+0.30}

\newcommand{\DNullSoneShort}{INCONC.}
\newcommand{\DNullSoneUp}{+0.58}

\newcommand{\DNullStwoEx}{+0.37}

\newcommand{\DNullStwoShort}{INCONC.}
\newcommand{\DNullStwoUp}{+0.64}
\newcommand{\DPairMaxCorr}{62}
\newcommand{\DPairMaxName}{R1 and R2}
\newcommand{\DPairMaxPct}{1.83}

\newcommand{\DPairMinCorr}{25}
\newcommand{\DPairMinName}{R1 and R3}
\newcommand{\DPairMinPct}{0.74}

\newcommand{\DPcCorrGsm}{1.39}
\newcommand{\DPcCorrHe}{0.61}

\newcommand{\DPcCorrMath}{5.33}
\newcommand{\DPcCorrMbpp}{1.56}
\newcommand{\DPcCorrMmlu}{0.41}
\newcommand{\DPcDacc}{\ensuremath{-}0.02}

\newcommand{\DPcSpeed}{1.18}

\newcommand{\DSoneCorr}{1.68}

\newcommand{\DSoneCorrExRLo}{+0.04}

\newcommand{\DSoneCorrGsm}{1.31}
\newcommand{\DSoneCorrHe}{0.41}

\newcommand{\DSoneCorrMath}{6.27}
\newcommand{\DSoneCorrMbpp}{1.43}
\newcommand{\DSoneCorrMmlu}{0.32}

\newcommand{\DStwoCorr}{1.76}

\newcommand{\DStwoCorrExRLo}{+0.11}

\newcommand{\DStwoCorrGsm}{1.59}
\newcommand{\DStwoCorrHe}{1.42}

\newcommand{\DStwoCorrMath}{5.87}
\newcommand{\DStwoCorrMbpp}{0.78}
\newcommand{\DStwoCorrMmlu}{0.41}

\newcommand{\ExCandAns}{13}
\newcommand{\ExRefAns}{26}
\newcommand{\ExSharedPct}{46}
\newcommand{\KoneExecDiff}{0}
\newcommand{\KoneExecProgs}{421}
\newcommand{\KoneRescoreDiff}{0}
\newcommand{\KoneRescoreRows}{10{,}140}
\newcommand{\KoneRescoreRuns}{3}

%% file: tab_default.tex
Reordered S1 & 1.68 & [1.31, 2.09] & +0.04 & \multicolumn{2}{c}{floor run} & +0.22 & [\ensuremath{-}0.09, +0.52] & 28.8 & INCONC. \\
Reordered S2 & 1.76 & [1.36, 2.15] & +0.11 & \multicolumn{2}{c}{floor run} & +0.22 & [\ensuremath{-}0.09, +0.53] & 29.2 & INCONC. \\
\midrule
Prefix caching & 1.62 & [1.26, 2.03] & \ensuremath{-}0.02 & \ensuremath{-}0.10 & [\ensuremath{-}0.36, +0.17] & \ensuremath{-}0.02 & [\ensuremath{-}0.34, +0.31] & 29.6 & INCONC. \\
Chunked prefill 512 & 1.60 & [1.25, 1.98] & \ensuremath{-}0.01 & \ensuremath{-}0.12 & [\ensuremath{-}0.38, +0.13] & +0.10 & [\ensuremath{-}0.20, +0.41] & 29.2 & PASS \\
Max 16 sequences & 1.47 & [1.13, 1.85] & \ensuremath{-}0.13 & \ensuremath{-}0.25 & [\ensuremath{-}0.52, +0.01] & \ensuremath{-}0.23 & [\ensuremath{-}0.52, +0.06] & 28.5 & PASS \\
No CUDA graphs & 1.77 & [1.39, 2.19] & +0.12 & +0.05 & [\ensuremath{-}0.22, +0.32] & +0.13 & [\ensuremath{-}0.19, +0.44] & 31.0 & INCONC. \\
Spec.\ n-gram & 1.78 & [1.39, 2.20] & +0.14 & +0.06 & [\ensuremath{-}0.20, +0.33] & \ensuremath{-}0.02 & [\ensuremath{-}0.33, +0.30] & 29.5 & INCONC. \\
Spec.\ EAGLE-3 & 1.63 & [1.27, 2.02] & \ensuremath{-}0.01 & \ensuremath{-}0.09 & [\ensuremath{-}0.33, +0.15] & +0.42 & [+0.12, +0.73] & 29.3 & INCONC. \\
Spec.\ draft model & 1.68 & [1.29, 2.07] & +0.05 & \ensuremath{-}0.04 & [\ensuremath{-}0.29, +0.21] & +0.28 & [\ensuremath{-}0.03, +0.59] & 29.0 & INCONC. \\
\midrule
FP8 KV cache & 3.36 & [2.80, 3.96] & +1.53 & +1.65 & [+1.19, +2.08] & +0.13 & [\ensuremath{-}0.37, +0.63] & 54.2 & FAIL \\
FP8 weights & 3.38 & [2.80, 4.01] & +1.53 & +1.67 & [+1.22, +2.11] & \ensuremath{-}0.29 & [\ensuremath{-}0.76, +0.19] & 55.4 & FAIL \\
\bottomrule

%% file: tab_bi.tex
Reordered S1 & 0.00 & [0.00, 0.00] & +0.00 & \multicolumn{2}{c}{floor run} & +0.00 & [+0.00, +0.00] & 0.0 & PASS \\
Reordered S2 & 0.00 & [0.00, 0.00] & +0.00 & \multicolumn{2}{c}{floor run} & +0.00 & [+0.00, +0.00] & 0.0 & PASS \\
\midrule
Prefix caching & 0.00 & [0.00, 0.00] & +0.00 & +0.00 & [+0.00, +0.00] & +0.00 & [+0.00, +0.00] & 0.0 & PASS \\
Chunked prefill 512 & 0.00 & [0.00, 0.00] & +0.00 & +0.00 & [+0.00, +0.00] & +0.00 & [+0.00, +0.00] & 0.0 & PASS \\
Max 16 sequences & 0.00 & [0.00, 0.00] & +0.00 & +0.00 & [+0.00, +0.00] & +0.00 & [+0.00, +0.00] & 0.0 & PASS \\
No CUDA graphs & 0.00 & [0.00, 0.00] & +0.00 & +0.00 & [+0.00, +0.00] & +0.00 & [+0.00, +0.00] & 0.0 & PASS \\
Spec.\ n-gram & 0.00 & [0.00, 0.00] & +0.00 & +0.00 & [+0.00, +0.00] & +0.00 & [+0.00, +0.00] & 0.0 & PASS \\
Spec.\ EAGLE-3 & 0.00 & [0.00, 0.00] & +0.00 & +0.00 & [+0.00, +0.00] & +0.00 & [+0.00, +0.00] & 0.0 & PASS \\
Spec.\ draft model & 0.00 & [0.00, 0.00] & +0.00 & +0.00 & [+0.00, +0.00] & +0.00 & [+0.00, +0.00] & 0.0 & PASS \\
\midrule
FP8 KV cache & 3.91 & [3.28, 4.53] & +3.37 & +3.91 & [+3.37, +4.43] & \ensuremath{-}0.47 & [\ensuremath{-}1.02, +0.11] & 58.2 & FAIL \\
FP8 weights & 3.43 & [2.83, 4.06] & +2.92 & +3.43 & [+2.92, +3.97] & \ensuremath{-}0.18 & [\ensuremath{-}0.70, +0.33] & 55.7 & FAIL \\
\bottomrule